\documentclass{article}
\usepackage{spconf,amsmath,graphicx}
\usepackage{amssymb}
\usepackage{subfigure}
\usepackage{subcaption}
\usepackage[hidelinks]{hyperref}
\usepackage{color}
\usepackage{siunitx}
\usepackage{booktabs}
\usepackage{etoolbox}
\usepackage{multirow}
\usepackage{amsmath}
\usepackage{changes}
\usepackage{bbm}

\newrobustcmd{\B}{\bfseries}

\title{Cross-Camera Human Motion Transfer by Time Series Analysis}
\name{ Yaping Zhao$^{1,2,3,*}$, Guanghan Li$^{2,*}$, Edmund Y. Lam$^{1,3,\dag}$}

\address{
$^1$Department of Electrical and Electronic Engineering, The University of Hong Kong, Hong Kong SAR \\
$^2$Tsinghua University, 30 Shuangqing Road, Beijing, China\\
$^3$ACCESS –- AI Chip Center for Emerging Smart Systems, Hong Kong SAR}
	
\begin{document}
\newcommand{\Amat}{{\bf A}}
\newcommand{\Bmat}{{\bf B}}
\newcommand{\Cmat}{{\bf C}}
\newcommand{\Dmat}{{\bf D}}
\newcommand{\Emat}[0]{{{\bf E}}}
\newcommand{\Fmat}[0]{{{\bf F}}}
\newcommand{\Gmat}[0]{{{\bf G}}}
\newcommand{\Hmat}[0]{{{\bf H}}}
\newcommand{\Imat}{{\bf I}}
\newcommand{\Jmat}[0]{{{\bf J}}}
\newcommand{\Kmat}[0]{{{\bf K}}}
\newcommand{\Lmat}[0]{{{\bf L}}}
\newcommand{\Mmat}[0]{{{\bf M}}}
\newcommand{\Nmat}[0]{{{\bf N}}}
\newcommand{\Omat}[0]{{{\bf O}}}
\newcommand{\Pmat}[0]{{{\bf P}}}
\newcommand{\Qmat}[0]{{{\bf Q}}}
\newcommand{\Rmat}[0]{{{\bf R}}}
\newcommand{\Smat}[0]{{{\bf S}}}
\newcommand{\Tmat}[0]{{{\bf T}}}
\newcommand{\Umat}{{{\bf U}}}
\newcommand{\Vmat}[0]{{{\bf V}}}
\newcommand{\Wmat}[0]{{{\bf W}}}
\newcommand{\Xmat}{{\bf X}}
\newcommand{\Ymat}[0]{{{\bf Y}}}
\newcommand{\Zmat}{{\bf Z}}

\newcommand{\av}{\boldsymbol{a}}
\newcommand{\Av}{\boldsymbol{A}}
\newcommand{\Cv}{\boldsymbol{C}}
\newcommand{\bv}{\boldsymbol{b}}
\newcommand{\cv}{{\boldsymbol{c}}}
\newcommand{\dv}{\boldsymbol{d}}
\newcommand{\ev}[0]{{\boldsymbol{e}}}
\newcommand{\fv}{\boldsymbol{f}}
\newcommand{\Fv}[0]{{\boldsymbol{F}}}
\newcommand{\gv}[0]{{\boldsymbol{g}}}
\newcommand{\hv}[0]{{\boldsymbol{h}}}
\newcommand{\iv}[0]{{\boldsymbol{i}}}
\newcommand{\jv}[0]{{\boldsymbol{j}}}
\newcommand{\kv}[0]{{\boldsymbol{k}}}
\newcommand{\lv}[0]{{\boldsymbol{l}}}
\newcommand{\mv}[0]{{\boldsymbol{m}}}
\newcommand{\nv}{\boldsymbol{n}}
\newcommand{\ov}[0]{{\boldsymbol{o}}}
\newcommand{\pv}[0]{{\boldsymbol{p}}}
\newcommand{\qv}[0]{{\boldsymbol{q}}}
\newcommand{\rv}[0]{{\boldsymbol{r}}}
\newcommand{\sv}[0]{{\boldsymbol{s}}}
\newcommand{\tv}[0]{{\boldsymbol{t}}}
\newcommand{\uv}[0]{{\boldsymbol{u}}}
\newcommand{\vv}{\boldsymbol{v}}
\newcommand{\wv}{\boldsymbol{w}}
\newcommand{\Wv}{\boldsymbol{W}}
\newcommand{\xv}{\boldsymbol{x}}
\newcommand{\yv}{\boldsymbol{y}}
\newcommand{\Xv}{\boldsymbol{X}}
\newcommand{\Yv}{\boldsymbol{Y}}
\newcommand{\zv}{\boldsymbol{z}}

\newcommand{\Gammamat}[0]{{\boldsymbol{\Gamma}}}
\newcommand{\Deltamat}[0]{{\boldsymbol{\Delta}}}
\newcommand{\Thetamat}{\boldsymbol{\Theta}}
\newcommand{\Lambdamat}{{\boldsymbol{\Lambda}}}
\newcommand{\Ximat}[0]{{\boldsymbol{\Xi}}}
\newcommand{\Pimat}[0]{{\boldsymbol{\Pi}} }
\newcommand{\Sigmamat}{\boldsymbol{\Sigma}}
\newcommand{\Upsilonmat}[0]{{\boldsymbol{\Upsilon}} }
\newcommand{\Phimat}{\boldsymbol{\Phi}}
\newcommand{\Psimat}{\boldsymbol{\Psi}}
\newcommand{\Omegamat}{{\boldsymbol{\Omega}}}

\newcommand{\Lambdav}{\bm{\Lambda}}
\newcommand{\alphav}{\boldsymbol{\alpha}}
\newcommand{\betav}[0]{{\boldsymbol{\beta}} }
\newcommand{\gammav}{{\boldsymbol{\gamma}}}
\newcommand{\deltav}[0]{{\boldsymbol{\delta}} }
\newcommand{\epsilonv}{\boldsymbol{\epsilon}}
\newcommand{\zetav}[0]{{\boldsymbol{\zeta}} }
\newcommand{\etav}[0]{{\boldsymbol{\eta}} }
\newcommand{\thetav}{\boldsymbol{\theta}}
\newcommand{\iotav}[0]{{\boldsymbol{\iota}} }
\newcommand{\kappav}{{\boldsymbol{\kappa}}}
\newcommand{\lambdav}[0]{{\boldsymbol{\lambda}} }
\newcommand{\muv}{\boldsymbol{\mu}}
\newcommand{\nuv}{{\boldsymbol{\nu}}}
\newcommand{\xiv}{{\boldsymbol{\xi}}}
\newcommand{\omicronv}[0]{{\boldsymbol{\omicron}} }
\newcommand{\piv}{\boldsymbol{\pi}}
\newcommand{\rhov}[0]{{\boldsymbol{\rho}} }
\newcommand{\sigmav}[0]{{\boldsymbol{\sigma}} }
\newcommand{\tauv}[0]{{\boldsymbol{\tau}} }
\newcommand{\upsilonv}[0]{{\boldsymbol{\upsilon}} }
\newcommand{\phiv}{\boldsymbol{\phi}}
\newcommand{\chiv}[0]{{\boldsymbol{\chi}} }
\newcommand{\psiv}{\boldsymbol{\psi}}
\newcommand{\omegav}[0]{{\boldsymbol{\omega}} }

\newcommand{\xin}[1]{{\textcolor{red}{#1}}}

\newcommand{\ts}{^{\top}}
\newcommand{\TV}{{\rm TV}}
\newtheorem{definition}{Definition}
\newtheorem{lemma}{Lemma}
\newtheorem{corollary}{Corollary}
\newtheorem{theorem}{Theorem}

\maketitle

\newcommand\blfootnote[1]{%
  \begingroup
  \renewcommand\thefootnote{}\footnote{#1}%
  \addtocounter{footnote}{-1}%
  \endgroup
}

\blfootnote{$^{*}$Equal contribution.  $^{\dag}$Corresponding author.}
\blfootnote{This work is supported in part by the Research Grants Council (GRF 17201822), by the Research Postgraduate Student Innovation Award (The University of Hong Kong), and by ACCESS –- AI Chip Center for Emerging Smart Systems, Hong Kong SAR.}

\vspace{-5mm}
\begin{abstract}
With advances in optical sensor technology, heterogeneous camera systems are increasingly used for high-resolution (HR) video acquisition and analysis. However, motion transfer across multiple cameras poses challenges. To address this, we propose an algorithm based on time series analysis that identifies motion seasonality and constructs an additive model to extract transferable patterns. Validated on real-world data, our algorithm demonstrates effectiveness and interpretability. Notably, it improves pose estimation in low-resolution videos by leveraging patterns derived from HR counterparts, enhancing practical utility. Code is available at: \href{https://github.com/IndigoPurple/TSAMT}{\textcolor{blue}{https://github.com/IndigoPurple/TSAMT}}. 

\end{abstract}


\section{introduction}

\begin{figure}
\includegraphics[width=\linewidth]{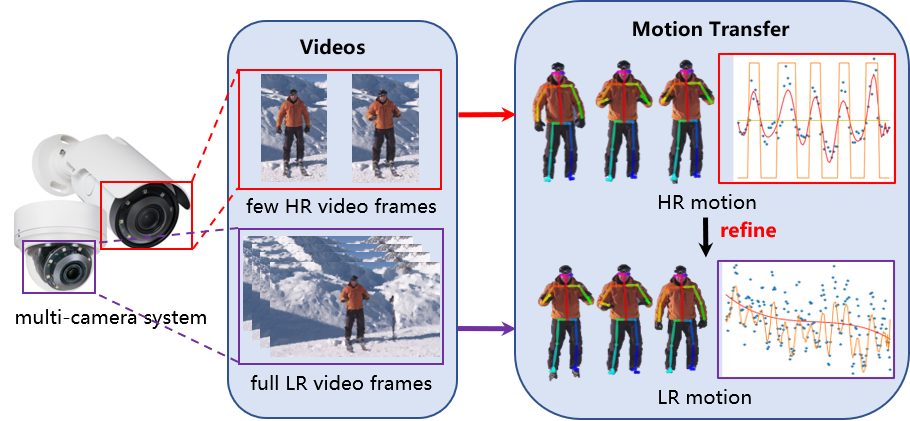}
\vspace{-15pt}
    \caption{
Our framework utilizes a multi-camera system to capture human-centric videos at multiple scales. By transfering extracted human motion patterns from high-resolution (HR) videos, we enhance pose estimation in low-resolution (LR) videos across different camera feeds. This approach ensures superior pose estimation irrespective of the video resolution.}
	\label{fig:teaser}
\end{figure}

Multi-camera systems, driven by advancements in optical sensors, enable computational imaging by combining short-focus and long-focus lenses. These systems capture videos with a wide field-of-view (FoV) and high-resolution (HR) local-view details. By infusing HR details into low-resolution (LR) videos, a balance is achieved between breadth and fine detail~\cite{giga_yuan2017multiscale, li2020zoom, zhao2021efenet}.
In scenarios like smart cities and live sports, multi-camera systems are widely used to capture multi-scale human-centric videos through hybrid-camera setups. However, transferring human movement data across different cameras, particularly with motion, poses a challenge. In this paper, we tackle this complex problem, illustrated in Figure~\ref{fig:teaser}.

\begin{figure}[]
    \centering
    \includegraphics[width=\linewidth]{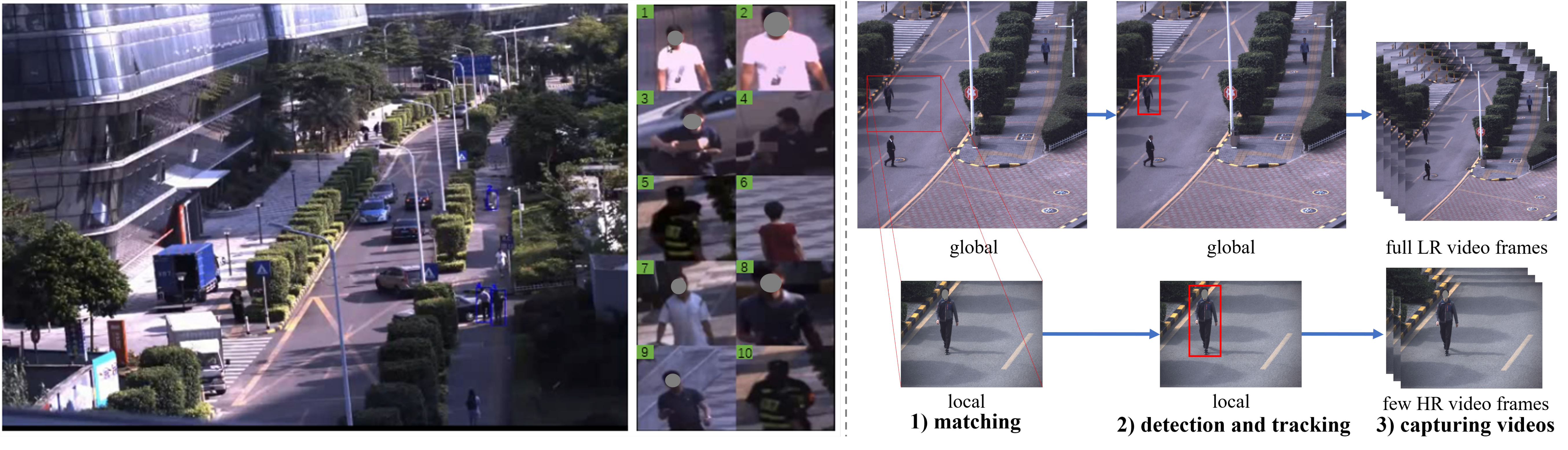}
    \vspace{-15pt}
    \caption{ For surveillance and security, a multi-camera system captures pedestrian trajectories in a large FoV video. Meanwhile, local-view cameras capture HR details of pedestrians.
     }
    
    \label{fig:monitor}
\end{figure}

Figure~\ref{fig:monitor} showcases a surveillance setup using a multi-camera system to track pedestrian movements across a large FoV video while capturing HR details with local-view cameras. However, integrating multi-scale human-centric videos into a composite video that preserves both a large FoV and HR detail presents challenges. Conventional algorithms struggle with nonrigid motion, subpar pose estimation on LR videos, and variations in camera settings, often resulting in blurry outputs. In contrast, deep learning approaches require intensive training data and time.

To address these challenges, we propose an efficient and interpretable motion transfer algorithm tailored for multi-camera systems. Our method, based on time series analysis, eliminates the need for training data and achieves accurate motion transfer. Experimental results on real-world data validate the effectiveness of our approach.

    Our main contributions are listed as follows:
    
    \begin{itemize}
        \item We introduce a motion transfer algorithm based on time series analysis for multi-camera systems, complete with a thorough and lucid mathematical formulation. 
        
        \item Our method is interpretable, avoids training procedures, and provides an algorithmic guarantee.

        \item Experiments on real-world data and downstream task substantiate the efficacy of our method.

    \end{itemize}

\section{related work}

\noindent\textbf{Synthesis of Human Motion.}
Prior research has focused on synthesizing human motion through statistical models and learned parameters~\cite{chai2007constraint, min2012motion, lau2009modeling}. However, these methods are limited by the variability in the training dataset, affecting the diversity of generated motion.

\noindent\textbf{Transfer of Human Motion.}
Deep learning approaches, such as pose-based prediction and image synthesis, have been applied to human motion transfer~\cite{villegas2017learning, villegas2018neural, ma2017pose, ma2018disentangled, siarohin2018deformable, chan2019everybody, liu2019liquid}. However, these methods heavily rely on training data and struggle with real-world data.

\noindent\textbf{Time Series Analysis.}
Time series analysis has found applications in various fields~\cite{cleveland1973analysis,brown2004smoothing,ghaderpour2018antileakage,ahmed2019introducing,ghaderpour2020least}, but its use in motion transfer between multiple cameras is lacking. We bridge this gap by introducing a motion transfer algorithm using time series analysis, with a focus on seasonality analysis~\cite{hylleberg1992modelling, wallis1974seasonal}.



\noindent\textbf{Multi-Camera Systems.}
Dual-camera systems, popular for their cost-efficiency, have been explored for image fusion and synthesis in super-resolution and denoising tasks~\cite{li2020zoom, zhao2022manet, jung2017enhancement, jang2020deep, zheng2018crossnet, tan2020crossnet++, zhao2021efenet,zhao2022cross,zhao2023improving}. 
Advanced imaging systems, such as snapshot compressive imaging systems~\cite{zhao2022deep,yang2022revisit,zhao2022mathematical,zhao2023deep,zhao2023transmission,zhenyuen2023solving,zhao2024sasa}, capture fast motion and may employ multiple cameras~\cite{wang2015dual,hauser2020dual}.
However, the challenge of efficient, effective, and explainable motion transfer remains unresolved. Our work addresses this challenge by proposing a motion transfer algorithm for multi-camera systems.

\section{Methodology}

In our multi-camera setup, we propose a time series analysis-based algorithm to enhance the LR pose sequence $\mathbf{\theta}^{L}$. We focus on seasonality analysis, leveraging the repetitive patterns observed in human actions like walking, running, and exercises in HR and LR videos.

Our motion transfer algorithm comprises five steps: (1) seasonality identification, (2) additive time series model construction, (3) periodic point detection, (4) additive factor extraction, and (5) motion pattern transfer. We demonstrate the process using $\theta_{1,1}$, representing the ankle joint's first axis-angle in the SMPL model~\cite{loper2015smpl}. The data is normalized to the range $[0,1]$ for analysis.

\begin{figure}[]
\centering
\subfigure[]{
\begin{minipage}[t]{0.48\linewidth}
\centering
\includegraphics[width=\linewidth]{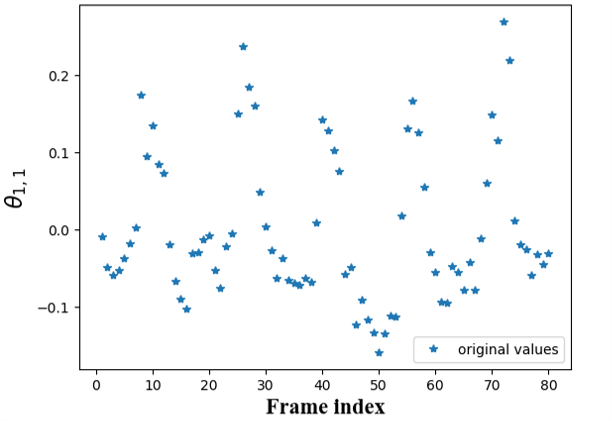}
\vspace{-20pt}
\label{fig:hr_or}
\end{minipage}%
}%
\subfigure[]{
\begin{minipage}[t]{0.52\linewidth}
\centering
\includegraphics[width=\linewidth]{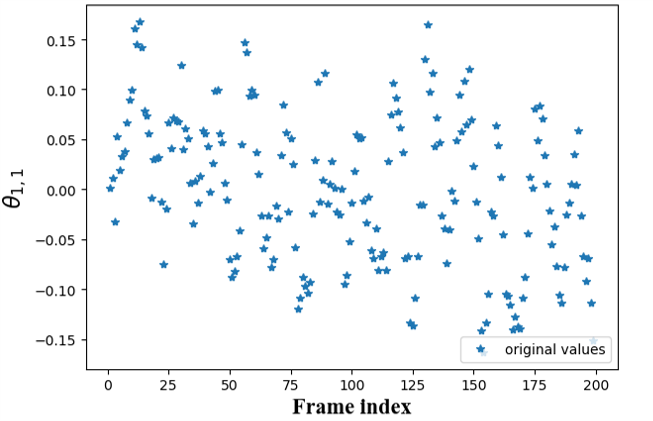}
\vspace{-20pt}
\label{fig:lr_or}
\end{minipage}%
}%
\vspace{-10pt}
\centering
\caption{ HR and LR motion data. (a) and (b) represent the $\theta_{1, 1}$ value of the HR and LR motion data, respectively.}
\end{figure}

Figure \ref{fig:hr_or} depicts an example of normalized $\theta_{1,1}$ values in a walking sequence with 80 HR frames, while Figure \ref{fig:lr_or} displays the corresponding values for 200 LR frames.

\begin{figure}[]
\centering
\subfigure[]{
\begin{minipage}[t]{0.5\linewidth}
\centering
\includegraphics[width=\linewidth]{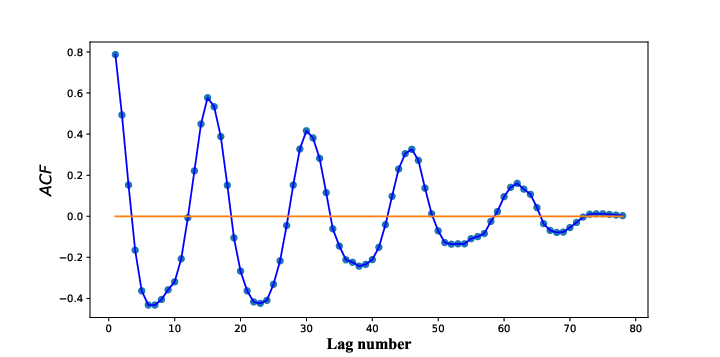}
\vspace{-15pt}
\label{fig:acf}
\end{minipage}%
}%
\subfigure[]{
\begin{minipage}[t]{0.5\linewidth}
\centering
\includegraphics[width=\linewidth]{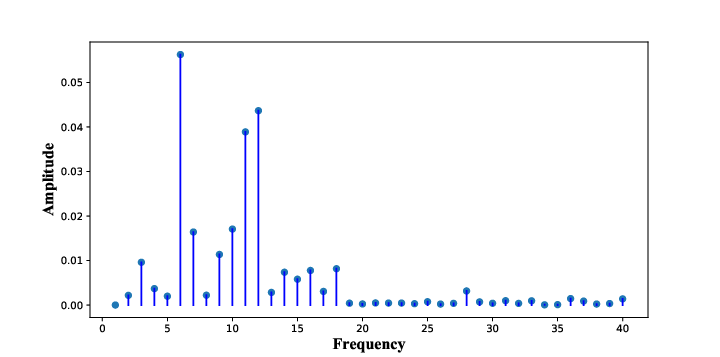}
\vspace{-15pt}
\label{fig:pyy}
\end{minipage}%
}%
\vspace{-10pt}
\centering
\caption{(a) Auto-correlation of HR motion data shows periodic variation with a maximum value during predefined intervals, gradually decreasing to zero.
(b) Fourier analysis of the HR motion data reveals a dominant response at $f = 5$, indicating a reference period of $16$ (derived from $80 / 5$).}
\end{figure}

\subsection{Identification of Seasonality}

We begin by utilizing the auto-correlation function (ACF)~\cite{plosser1979short} to identify the cyclical nature of HR motion data. Figure \ref{fig:acf} illustrates the periodic variations of the ACF curve, reaching peak values at predefined intervals and decaying to zero.

Although the cycle length of human motion is not fixed, it typically revolves around a constant value known as the "reference period" ($l$). Using Fourier analysis, we estimate $l$ based on the relationship:

\begin{equation}
l = n / f,
\end{equation}

where $n$ represents the total number of frames in the motion sequence, and $f$ corresponds to the frequency with the strongest response. Figure \ref{fig:pyy} displays the Fourier series of the HR motion data. In this case, the frequency $f = 5$ exhibits the strongest response, resulting in a reference period of $16$ frames ($80 / 5$).

\begin{figure}[]
\centering
\subfigure[]{
\begin{minipage}[t]{\linewidth}
\centering
\includegraphics[width=0.7\linewidth]{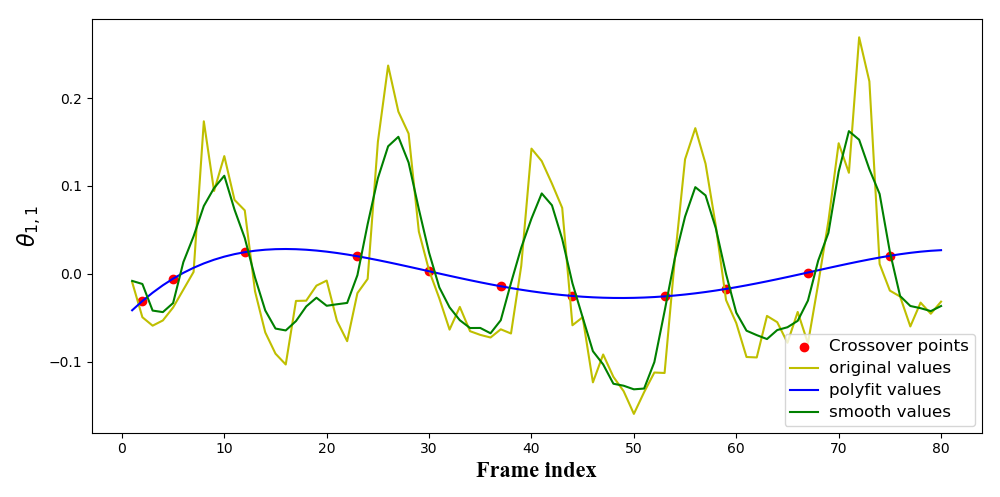}
\vspace{-25pt}
\label{fig:crossover}
\end{minipage}%
}%
\\
\subfigure[]{
\begin{minipage}[t]{0.55\linewidth}
\centering
\includegraphics[width=\linewidth]{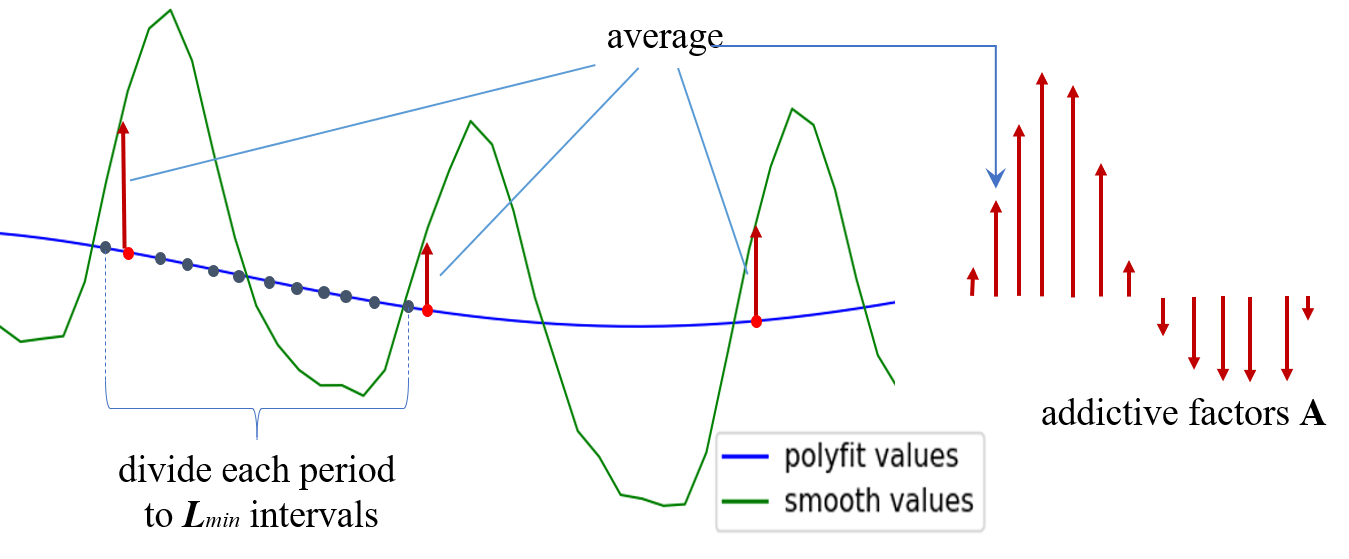}
\vspace{-25pt}
\label{fig:motion-refine1}
\end{minipage}%
}%
\subfigure[]{
\begin{minipage}[t]{0.45\linewidth}
\centering
\includegraphics[width=\linewidth]{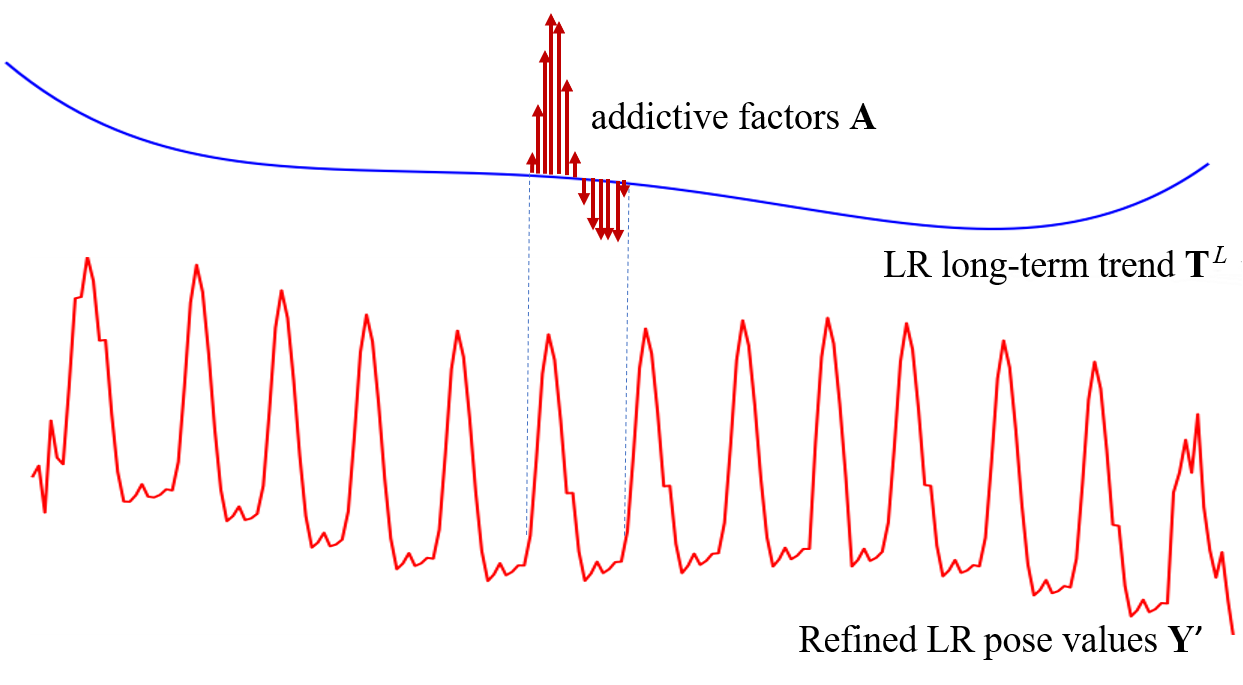}
\vspace{-25pt}
\label{fig:motion-refine2}
\end{minipage}%
}%
\vspace{-10pt}
\centering
\caption{ (a) Crossover points of the blue and dark green curves indicate the locations of the periods, marked as red circles.
(b) To refine LR pose values using HR ones, we estimate the additive factor $\mathbf{A}$ by averaging all the periods.
(c) By adding the additive factor to the long-term trend $\mathbf{T}^L$, we generate the final poses for refining LR pose values with HR ones.}
\end{figure}

\subsection{Construction of Additive Time Series Model}

To deconstruct the LR pose values, we employ an additive time-series model~\cite{maravall1987prototypical}. The decomposition is expressed as:
\begin{equation}
    \mathbf{Y} = \mathbf{S} + \mathbf{T} + \mathbf{E},
    \label{eq:ts-full}
\end{equation}
where $\mathbf{Y}$ represents the original pose values, $\mathbf{S}$ captures short-term variations, $\mathbf{T}$ models the long-term trend, and $\mathbf{E}$ accounts for noise. We estimate the long-term trend of the LR sequence, $\mathbf{T}^L$, by polynomial fitting using the least squares method:
\begin{equation}
    \mathbf{T}^L = c_0 + c_1 \theta_{1,1}^L + c_2 {(\theta_{1,1}^L)}^{2} + \cdots + c_{n-1} {(\theta_{1,1}^L)}^{n-1} + c_{n} {(\theta_{1,1}^L)}^n,
\end{equation}
where $c_0, \cdots, c_n$ are constant parameters and $\theta_{1,1}^L$ represents the $\theta_{1, 1}$ value of the LR data. The order $n$ of the fitting function is chosen to satisfy $1^{-10} < |c_n| < f$, with $c_n$ as the coefficient of the highest-order term and $f$ denoting the number of periods in a pose value sequence.

\subsection{Location of Periodic Points}

In the next phase, we employ a moving average to smooth pose values and identify period crossover points (red circles in Figure~\ref{fig:crossover}). Period indices within the range $[0.8l, 1.2l]$ are retained, and their averages yield additive factors $\mathbf{A}$.

By adding $\mathbf{A}$ to the long-term trend $\mathbf{T}$, we obtain refined LR poses (Figure \ref{fig:motion-refine1} and Figure \ref{fig:motion-refine2}). To identify frame indices corresponding to crossover points, we detect positive and negative sign changes in the difference between the smoothed pose curve and the trend curve. Imperfections in polynomial fitting are addressed by validating neighboring periodic indices. For each periodic index $p$, we search for a neighboring index $p'$ satisfying:
\begin{equation}
\begin{split}
    p' = &\mathop{\arg\min}_{p'} \left| p' - p \right|, \\
    &\mathrm{s.t.} \left| \left| p' - p \right| - l \right| < (1 - \alpha) \times l,
\end{split}
\end{equation}
Here, we set the confidence level $\alpha$ to 80\%, with $l$ as the reference period. This validation process ensures neighboring periodic points fall within the confidence interval.

\subsection{Extraction of Additive Factor}

After identifying the periods in the time series, we calculate the minimum period number $l_{min}$ for both HR and LR sequences and divide each period into $l_{min}$ intervals uniformly. To extract the additive factor for motion pattern transfer, we subtract the long-term trend component from the HR pose values using Equation~\ref{eq:ts-full}:
\begin{equation}
    \mathbf{A} = \mathbf{Y}^H - \mathbf{T}^H = \mathbf{S}^H + \mathbf{E}^H,
    \label{eq:a-definition}
\end{equation}
Here, $\mathbf{A}$ represents the additive factor, composed of short-term variation $\mathbf{S}^H$ and noise component $\mathbf{E}^H$.

Equation~\ref{eq:a-definition} enables the extraction of the periodically repetitive pattern $\mathbf{A}$ for motion transfer (Figure~\ref{fig:motion-refine1}). To achieve comprehensive motion transfer, we compute the mean value across all periods as the mean additive factor $\overline{\mathbf{A}} = (\overline{a}_{1},..., \overline{a}_{i},..., \overline{a}_{l})$:
\begin{equation}
    \overline{a}_{j} = \frac{\sum_{i = 1}^{n} \mathbbm{1}\{ \phi(i) = j \} a_{i}}{(f \times m)}, \qquad a_{i} \in \mathbf{A},
    \label{eq:a-mean}
\end{equation}
In Equation~\ref{eq:a-mean}, $a_{i}$ refers to the $i_{th}$ value of $\mathbf{A}$ in the $i_{th}$ interval of a period, $f$ is the number of periods, $m$ is the number of values in the $i_{th}$ interval, and $\phi$ is a correspondence function that maps the frame index $i$ to the $j_{th}$ interval of a period:
\begin{equation}
    \phi(i) = j.
    \label{eq:phi}
\end{equation}


\subsection{Pattern Transfer}

In the final phase, we integrate the additive factors $\overline{\mathbf{A}}$ with the LR long-term trend $T^L$, aligning it with the extracted regular patterns from the HR motion sequence (Figure~\ref{fig:motion-refine2}). The optimized LR pose values $\mathbf{Y'} = (y'_{1},..., y'_{i},...,y'_{n})$ are computed using the formula:
\begin{equation}
    {y'}_{i} = t_{i} + \overline{a}_{\phi(i)},
\end{equation}
Here, ${y'}_{i}$ represents the i$_{th}$ value of the enhanced LR pose, and $t_{i}$ is the $i_{th}$ value of the LR long-term trend $\mathbf{T}^L$. This pattern transfer approach yields a refined representation of the original LR poses, providing a higher quality approximation of the initial HR sequence.

\section{Experiment}


To validate our proposed algorithm's effectiveness, we apply it to real-world data from a dual-camera system~\cite{li2020zoom}. Using a limited number of HR frames and a complete series of LR frames, we initially estimate poses using OpenPose~\cite{cao2019openpose}, resulting in pose sequences.

\begin{figure}[!htb]
\centering
\subfigure[]{
\begin{minipage}[t]{0.5\linewidth}
\centering
\includegraphics[width=\linewidth]{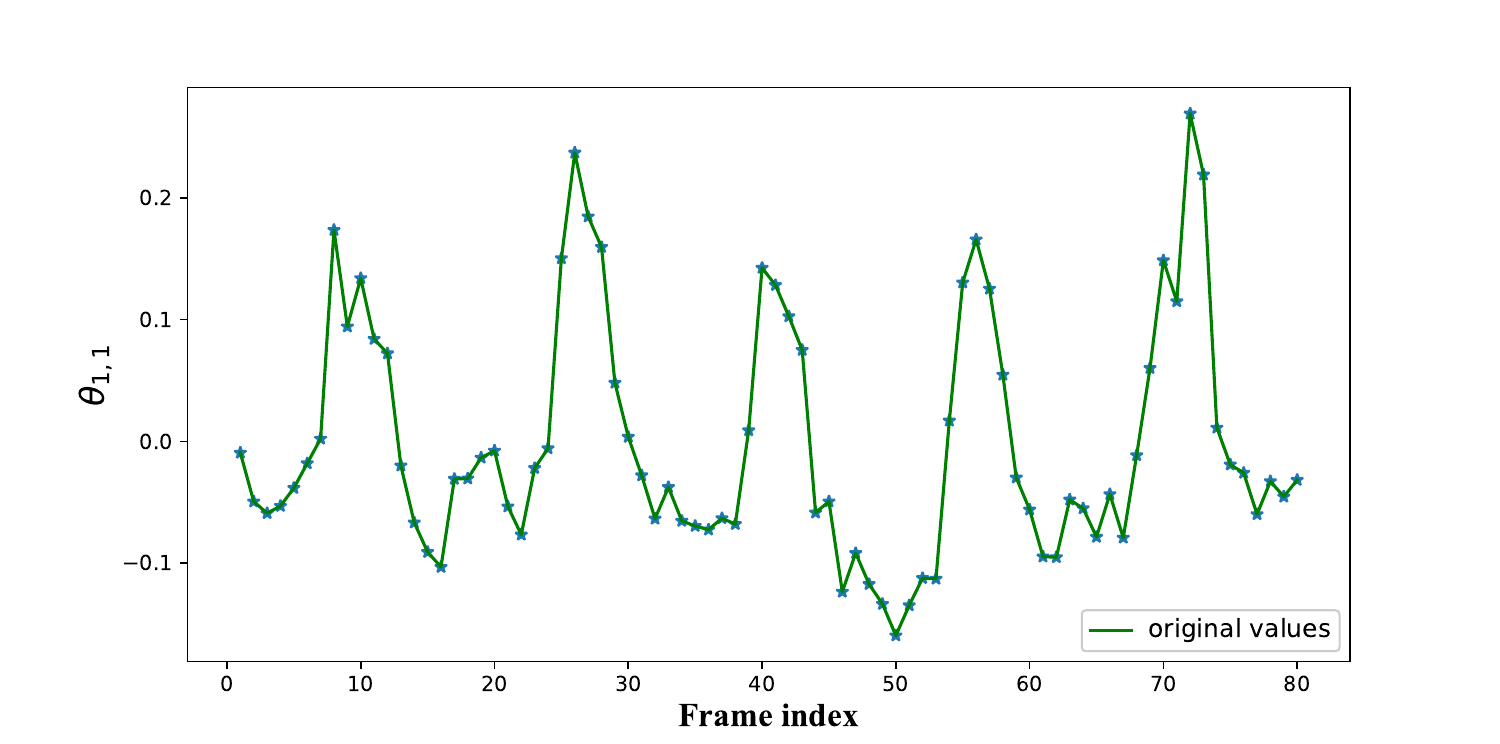}
\vspace{-25pt}
\label{fig:hr_ex}
\end{minipage}%
}%
\subfigure[]{
\begin{minipage}[t]{0.5\linewidth}
\centering
\includegraphics[width=\linewidth]{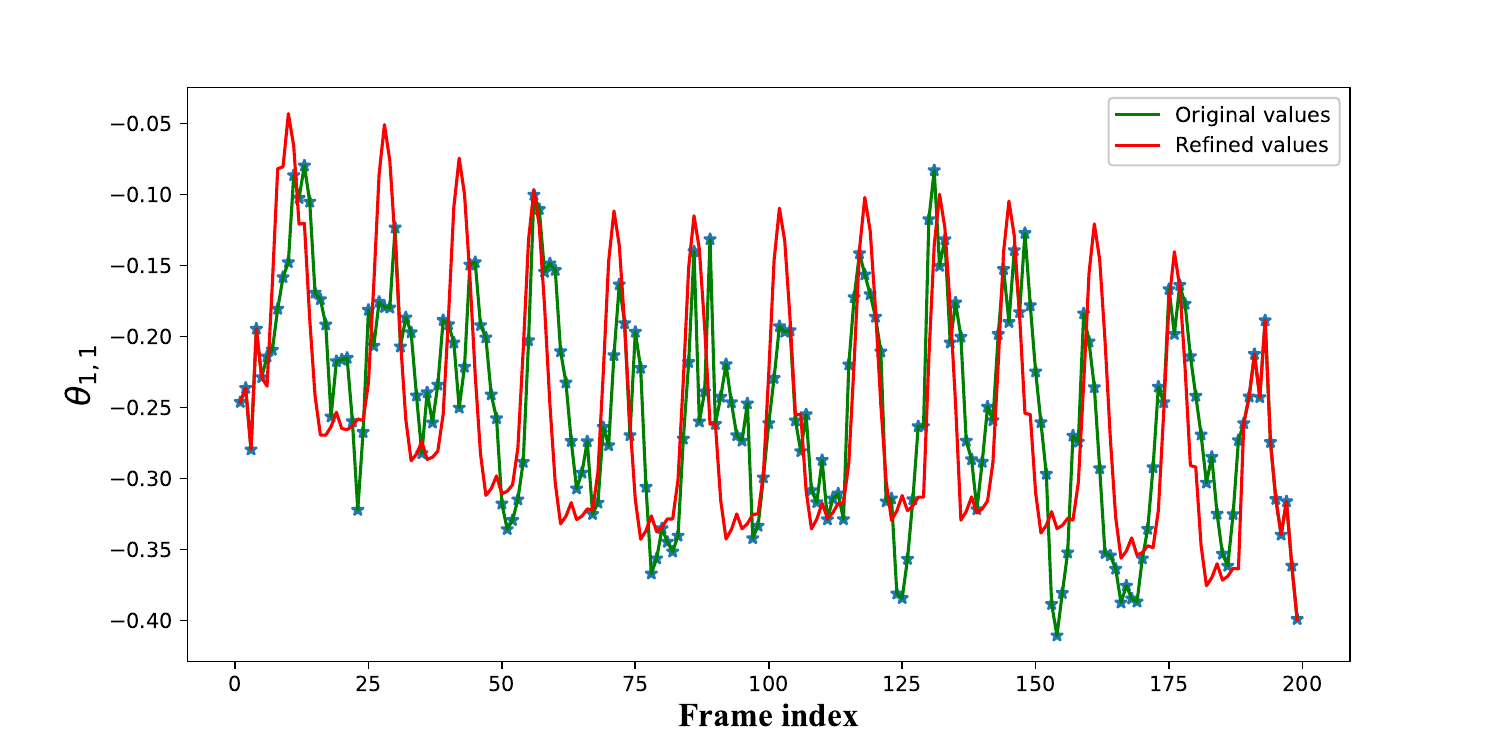}
\vspace{-25pt}
\label{fig:lr_ex}
\end{minipage}%
}%
\vspace{-10pt}
\centering
\caption{ HR and LR motion data. (a) and (b) represent the $\theta_{1, 1}$ value of the HR and LR motion data, respectively.}
\label{fig:exp1}
\end{figure}

\begin{figure}[!htb]
\centering
\subfigure[]{
\begin{minipage}[!h]{\linewidth}
\centering
\includegraphics[width=\linewidth]{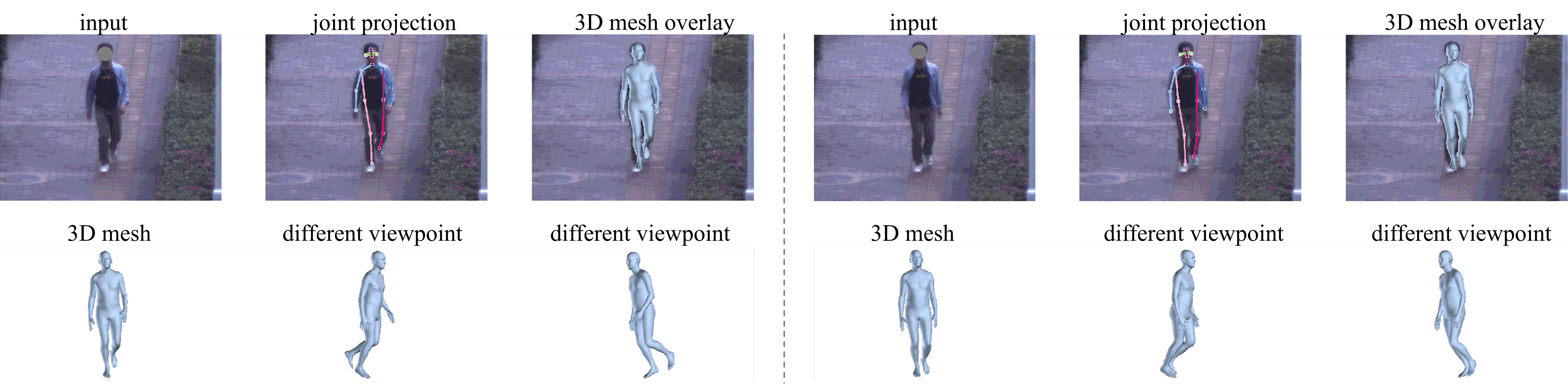}
\vspace{-15pt}
\label{fig:hmr1}
\end{minipage}%
}%
\\
\subfigure[]{
\begin{minipage}[!h]{\linewidth}
\centering
\includegraphics[width=\linewidth]{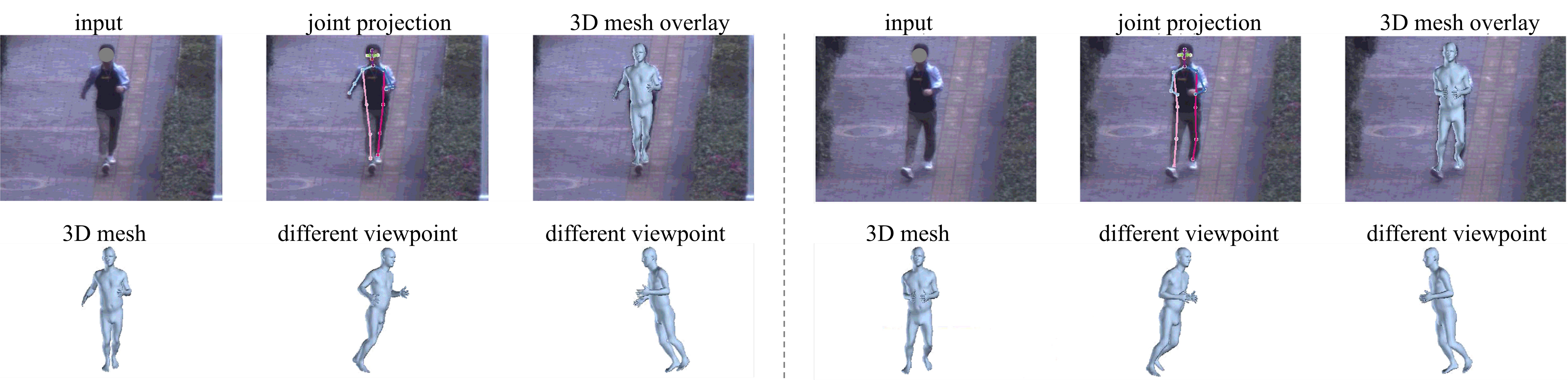}
\vspace{-15pt}
\label{fig:hmr2}
\end{minipage}%
}%
\vspace{-10pt}
\centering
\caption{The SMPL model~\cite{smpl_loper2015smpl} is adopted to represent 3D body and the HMR~\cite{kanazawa2018end} is performed to reconstruct 3D human meshes from 2D video frames. (a) and (b) show successful cases and failure cases, respectively.}
\end{figure}

\begin{figure}
    \centering
    \includegraphics[width=\linewidth]{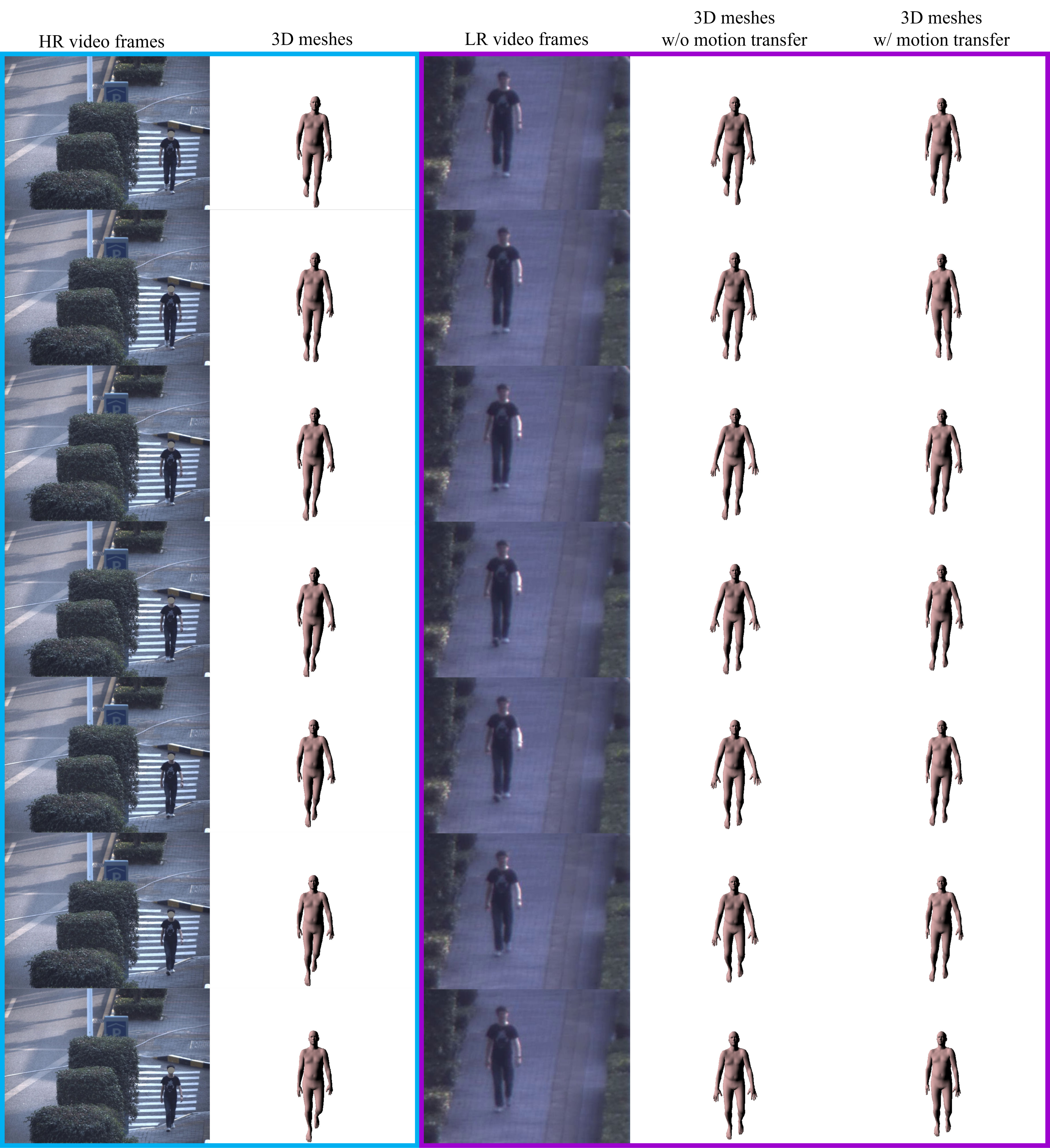}
    \caption{Our motion transfer technique improves 3D human mesh reconstruction performance. In the LR video, we observed misaligned arm positions. However, the results with motion transfer effectively correct the human poses.}
    \label{fig:hmr_motion}
\end{figure}

\begin{figure}[!h]
    \includegraphics[width = 1\linewidth]{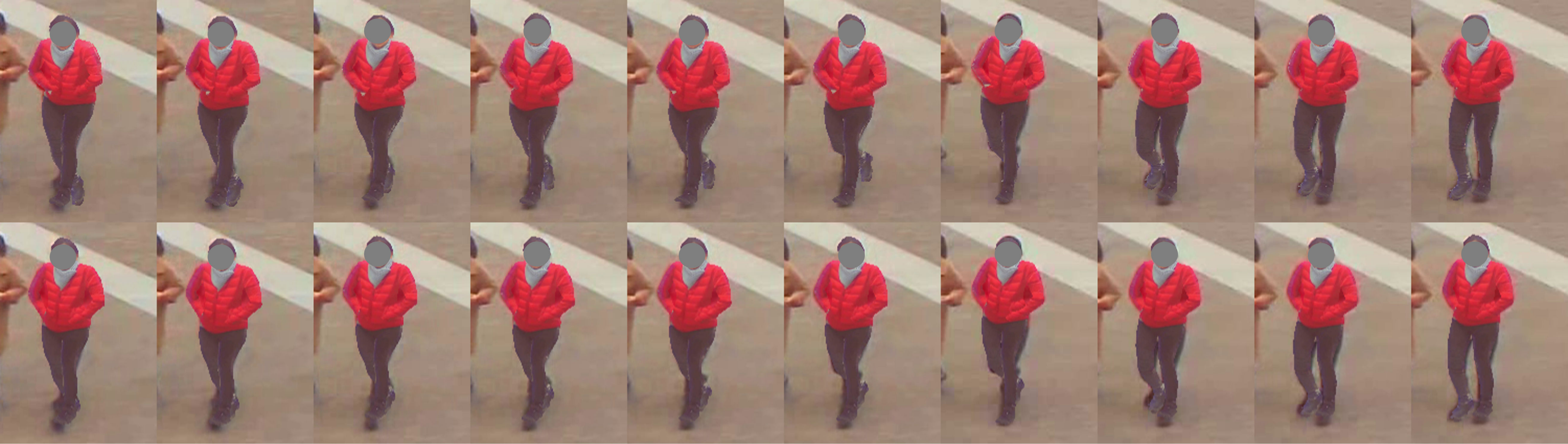}
    \caption{Performance on synthesizing human details across heterogeneous cameras. Top: without our motion transfer method, jitters are observed on the legs. Bottom: with ours. }
    \label{fig:motion}
\end{figure}

In Figure~\ref{fig:exp1}, the HR pose sequence is of superior quality (green curves), while the LR one suffers from severe noise. Our motion transfer approach successfully refines the LR pose values, as shown in Figure~\ref{fig:lr_ex}. Our results preserve the long-term correlation in the LR sequence and enhance its quality by transferring motion patterns from the HR sequence.

For 3D body representation, we employ the SMPL model~\cite{smpl_loper2015smpl} and use the HMR methodology~\cite{kanazawa2018end} to reconstruct 3D human meshes from 2D frames. While satisfactory results are achieved in straightforward cases (e.g., minimal leg movement, arms close to knees; Figure~\ref{fig:hmr1}), HMR struggles with accurate estimation when the subject walks briskly and swings their arms vigorously (Figure~\ref{fig:hmr2}).

To enhance 3D mesh reconstruction and demonstrate the efficacy of our motion transfer algorithm, we apply it to HR and LR real-world videos of a pedestrian walking. Without our method, the reconstructed 3D human meshes exhibit inaccuracies and anomalies due to the LR video quality (Figure~\ref{fig:hmr_motion}). However, our method corrects these human poses.

Furthermore, we integrate the motion transfer method into a downstream vision task of embedding HR human details into LR videos. Without our algorithm, direct synthesis of human details is inaccurate, resulting in jittering and unnatural poses (Figure~\ref{fig:motion}). However, applying motion transfer reduces motion jittering, demonstrating the utility of our approach.


Given the unavailability of ground truths of human body joints in real-world scenarios, direct quantitative evaluations are infeasible. However, by seamlessly integrating our approach with downstream tasks, we facilitate quantitative assessments. To this end, we employ video super-resolution with texture transfer as a downstream task to quantitatively evaluate the effectiveness of our approach. For detailed experimental results, please refer to the supplementary material, which is provided due to space constraints:
\\\href{https://github.com/IndigoPurple/TSAMT/blob/main/sm/sm.pdf}{\textcolor{blue}{github.com/IndigoPurple/TSAMT/blob/main/sm/sm.pdf}}.



\section{Conclusion}

We present an algorithm for motion transfer among multiple cameras, based on time series analysis. Our approach has a comprehensive and well-defined mathematical formulation and does not necessitate training data, thereby ensuring efficiency and interpretability. Through rigorous evaluations on real-world data, we showcase the effectiveness of our method. 

\small
{\bibliographystyle{IEEEtran}
\bibliography{refs_motion}
}
\end{document}